\documentclass[conference]{IEEEtran}
	  \usepackage{cite}
		\ifCLASSINFOpdf
		\else
		\fi
		\usepackage{ucs}
		\usepackage{amsmath}
		\usepackage{amsfonts}
		\usepackage{amssymb}
		\usepackage{makeidx}
		\usepackage{breqn} %necessario para usar o ambiente 'dmath'
		\usepackage{url}
		\usepackage{hyperref}
		 \usepackage{listings}
		 \usepackage{color}
		 
		 \definecolor{dkgreen}{rgb}{0,0.6,0}
		 \definecolor{gray}{rgb}{0.5,0.5,0.5}
		 \definecolor{mauve}{rgb}{0.58,0,0.82}
		 
		\usepackage{graphicx}
		
\begin{document}
		%
		% paper title
		% Titles are generally capitalized except for words such as a, an, and, as,
		% at, but, by, for, in, nor, of, on, or, the, to and up, which are usually
		% not capitalized unless they are the first or last word of the title.
		% Linebreaks \\ can be used within to get better formatting as desired.
		% Do not put math or special symbols in the title.
	%	\title{Designing Smart Things using\\ Evolutionary Neural Networks} %Embodied Agents}
	
	%\title{Automatic creation of Smart Things through Evolutionary Neural Networks}
	
%	\title{Automatic creation of Smart Things' Controllers through Evolutionary Neural Networks}
	%\title{Evolution of Neural Controllers for \\Smart Things}
	
	%\title{Prototyping Smart Things with \\Evolved Neural Networks}
	%	\title{Engineering Smart Things with \\Evolved Neural Networks}	
	
%	\title{Engineering Smart Things based on \\Evolved Networks}	
%	\title{Engineering Cognitive Embedded Systems}
		\title{Engineering Cooperative Smart Things \\based on Embodied Cognition}	
	
%	\title{Engineering 	Awareness Agents based on \\Evolved Network}	
	
		%\title{Engineering Autonomous Things with \\Evolved Neural Networks}	
	%	\title{An Approach for Fog Computing : \\Engineering Smart Things with \\Evolved Neural Networks}
	%	\title{Designing Cooperative Smart Things using \\Evolutionary Embodied Agents} 
		
		% author names and affiliations
		% use a multiple column layout for up to three different
		% affiliations
		\author{\IEEEauthorblockN{Nathalia Moraes do Nascimento\\ and Carlos Jos{\'e} Pereira de Lucena}
%	%\author{\IEEEauthorblockN{XXX\\ and XXX}
		\IEEEauthorblockA{Software Engineering Laboratory (LES), Department of Informatics\\
		Pontifical Catholic University of Rio de Janeiro\\
		Rio de Janeiro, Brazil\\
		Email: {nnascimento, lucena}@inf.puc-rio.br}
%		\and
%			\IEEEauthorblockN{XXX}
%					\IEEEauthorblockA{XXX, Department of Informatics\\
%						 University\\
%						city, Country\\
%					Email: {,}@xxx}
%		\IEEEauthorblockN{Alexandre Meslin\\ and Noemi Rodriguez}
%		\IEEEauthorblockA{Sensor Network Lab, Department of Informatics\\
%			Pontifical Catholic University of Rio de Janeiro\\
%			Rio de Janeiro, Brazil\\
%		Email: {,}@inf.puc-rio.br}
		}
		
		% conference papers do not typically use \thanks and this command
		% is locked out in conference mode. If really needed, such as for
		% the acknowledgment of grants, issue a \IEEEoverridecommandlockouts
		% after \documentclass
		
		% for over three affiliations, or if they all won't fit within the width
		% of the page, use this alternative format:
		% 
		%\author{\IEEEauthorblockN{Michael Shell\IEEEauthorrefmark{1},
		%Homer Simpson\IEEEauthorrefmark{2},
		%James Kirk\IEEEauthorrefmark{3}, 
		%Montgomery Scott\IEEEauthorrefmark{3} and
		%Eldon Tyrell\IEEEauthorrefmark{4}}
		%\IEEEauthorblockA{\IEEEauthorrefmark{1}School of Electrical and Computer Engineering\\
		%Georgia Institute of Technology,
		%Atlanta, Georgia 30332--0250\\ Email: see http://www.michaelshell.org/contact.html}
		%\IEEEauthorblockA{\IEEEauthorrefmark{2}Twentieth Century Fox, Springfield, USA\\
		%Email: homer@thesimpsons.com}
		%\IEEEauthorblockA{\IEEEauthorrefmark{3}Starfleet Academy, San Francisco, California 96678-2391\\
		%Telephone: (800) 555--1212, Fax: (888) 555--1212}
		%\IEEEauthorblockA{\IEEEauthorrefmark{4}Tyrell Inc., 123 Replicant Street, Los Angeles, California 90210--4321}}

		% use for special paper notices
		%\IEEEspecialpapernotice{(Invited Paper)}

		% make the title area
		\maketitle
		
		% As a general rule, do not put math, special symbols or citations
		% in the abstract
	%	\begin{abstract}
%		Learning as an adaptive process.
%		From centralized approaches for distributed processing
%		Fog Computing
%		\end{abstract}
		
		\begin{abstract}
			
				The goal of the Internet of Things (IoT) is to transform any thing around us, such as a trash can or a street light, into a smart thing. A smart thing has the ability of sensing, processing, communicating and/or actuating. In order to achieve the goal of a smart IoT application, such as minimizing waste transportation costs or reducing energy consumption, the smart things in the application scenario must cooperate with each other without a centralized control. Inspired by known approaches to design swarm of cooperative and autonomous robots, we modeled our smart things based on the embodied cognition concept. Each smart thing is a physical agent with a body composed of a microcontroller, sensors and actuators, and a brain that is represented by an artificial neural network. This type of agent is commonly called an embodied agent. The behavior of these embodied agents is autonomously configured through an evolutionary algorithm that is triggered according to the application performance. To illustrate, we have designed three homogeneous prototypes for smart street lights based on an evolved network. This application has shown that the proposed approach results in a feasible way of modeling decentralized smart things with self-developed and cooperative capabilities.

\textit{Keywords-embodied cognition; cognitive system; cognitive embedded systems; evolved network; neural network; multiagent system; smart things; internet of things; cooperative systems; self-developed systems;emergent communication system}
		\end{abstract}
		% no keywords

		% For peer review papers, you can put extra information on the cover
		% page as needed:
		% \ifCLASSOPTIONpeerreview
		% \begin{center} \bfseries EDICS Category: 3-BBND \end{center}
		% \fi
		%
		% For peerreview papers, this IEEEtran command inserts a page break and
		% creates the second title. It will be ignored for other modes.
		\IEEEpeerreviewmaketitle

		\section{Introduction}
		
		A few years ago, Kephart and Chess (2003) \cite{kephart2003vision} called the global goal to connect trillions of computing devices to the Internet the nightmare of ubiquitous computing. The reason for that is that to reach this global goal requires a lot of skilled Information Technology (IT) professionals to create millions of lines of code, and install, configure, tune, and maintain these devices. According to Kephart (2005) \cite{kephart2005research}, in some years, IT environments will be impossible to be administered, even by the most skilled IT professionals.

		Predicting the emergence of this problem, in 2001 the IBM company suggested the creation of autonomic computing \cite{horn2001autonomic}.  IBM recognized that the only viable solution to resolve this problem was to endow systems and the components that comprise them with the ability to manage themselves in accordance with high-level  objectives specified by humans \cite{kephart2005research}. Therefore, IBM proposed systems with self-developed capabilities.  The company emphasized the need of automating IT key tasks, such as coding, configuring, and maintaining systems, based on the progress observed in the automation of manual tasks in agriculture.

		Other IT companies agreed with IBM and then generated their own proposals \cite{hp,microsoft}. However, the IT Industry interest in the development of self-management devices is not yet very evident. As a result, not only the goal of the Internet of Things (IoT) to connect billions of devices to the Internet has not been reached, but also we have been experiencing the problems previously listed by Kephart and Chess  (2003) \cite{kephart2003vision}. %IBM in \cite{horn2001autonomic}.  
		In fact, there is a lack of software to support the development of a huge number of different IoT applications.
		
		%The truth is that companies and researchers are so busy competing to define the official protocol and architecture for the IoT, that very little research has been done to provide a sophisticated control to manage all these billions of things. As a result, there is a lack of software to support the development of a huge number of different IoT applications.

		In this context, we have been investigating how to create applications based on the IoT with self-developed and cooperative capabilities. To this end, our approach consists in:
		\begin{itemize}
			\item Developing smart things: %Autonomous Things
			\begin{itemize}
				\item Things that are autonomous and able to execute complex behavior without the need for centralized control to manage their interaction.
				\item Things that are able to have behavior assigned at design-time and/or at run-time.
			\end{itemize}
			\item Providing mechanisms to allow things to self-adapt, improve their own behavior and cooperate;
		\end{itemize}
	%	\section{Out of Scope}
	
	To reach these objectives, we previously developed a generic software basis for IoT, which is called the ``Framework for Internet of Things" (FIoT) \cite{do2017fiot}.  The framework approach was used to develop the common requirements among IoT applications and implement a reusable architecture \cite{Markiewicz:2001:OOF:372765.372771}. We developed FIoT according to the following directions:	
		
		\begin{enumerate}
				\item To create autonomous things and a distributed control, we modeled the framework based on a multiagent approach \cite{oldPaperSMA}. According to Cetnarowicz et. al (1996) \cite{oldPaperSMA}, the active agent was invented as a basic element from which distributed and decentralized systems could be built. In our approach, we considered the use of embodied agents, which is typically used to model and control autonomous physical objects that are situated in actual environments \cite{ecagents}.
				\item To control the things, we chose a control architecture based on artificial neural networks  \cite{haykin1994neural}. A neural network is a well known approach to dynamically provide responses and automatically create a mapping of input-output relations  \cite{haykin1994neural}. In addition, it is commonly used as an internal controller of embodied agents \cite{maroccoNolfi}.%,Nolfi97learningto}.
				\item To make things self-adaptive, we proposed the use of the IBM control-loop \cite{jacob2004practical} combined with various Machine Learning (ML) techniques, notably supervised learning and evolutionary algorithms \cite{FloreanoBook}.

		\end{enumerate}

As the development of smart things is part of a broader context, a set of related aspects will be left out of the scope of this work. Thus, the following approaches are not directly addressed by this work: security, ontology, protocols and scalability.

%evolved networks		
%SE nao focar no framework...
%The goal of this paper is to show how to engineer smart things with self-developed and cooperative capabilities based on evolved networks.  		
The goal of this paper is to show how FIoT can be used to prototype physical smart things based on embodied cognition. 
Previously, we modeled and simulated smart traffic lights in \cite{do2017fiot}, which were tested in a simulated car traffic application. 
%We presented a simulated car traffic application, where each traffic light in the simulation had a micro-controller that was used to calculate the number of vehicles per time period, interact with the closest segment, and change the traffic light. 
However, we only provided simulated smart things. Therefore, we did not show how to transfer the evolved controller to physical smart things. For instance, we created a simpler experiment, but we will show all the steps of engineering smart things using evolved neural networks. We will present the steps of modeling and evolving a neural network in a simulated environment, and the step of transferring this evolved network to physical devices.

 We present this experiment in Section \ref{sec:lights}. 
 It presents the experimental setup, results, and evaluation. The remainder of this paper is organized as follows. Section \ref{sec:related} presents the related work. Section  \ref{sec:background}  describes the background for the proposed approach. %Section \ref{sec:challenge} discusses the challenges of implementing this idea in a real world. 
 The paper ends with conclusive remarks in Section \ref{sec:conclusion}.

		\section{Related Work} \label{sec:related}
	
%Most of IoT applications aggregate IoT data on the cloud for intelligent processing to help make decisions and take actions. According to  Sheth \cite{sheth2016internet}, in order to avoid the daily generation of billions of unused data, new technologies must be integrated to allow the analyzes of data in real time. One solution is to downscale this processing to the edges \cite{sheth2016internet}. Therefore, each IoT component must be smart.%able to process data and take decisions. 

%Smart Things (or Smart Objects) are autonomous objects connected to the Internet, which can be identified by systems and, possibly, communicate with other devices. These objects have some hardware components, such as radio for communication, CPU to process tasks, sensors and actuators. Also, a smart thing can interpret the information created within themselves and around the neighboring external world where they are situated \cite{bookSmartObjects} \cite{smartThings}. 

There are some research results in the literature about smart things, which use a kind of self-developed approach \cite{katasonov2008smart,baresi2014short,zhu2014minson,de2016intelligent}. Baresi et al. \cite{baresi2014short}, for example, provide a simulation for a smart green house scenario. In their simulation, flowers are distributed in different rooms based on specific characteristics. If a flower is sick, it will be allocated to another room or its room's configuration will change. For this purpose, the authors use adaptive techniques to perform discovery, self-configuration, and communication among heterogeneous things. However, most of this research presents only simulated smart things and does not show how to transfer their approaches from a simulated smart thing to a physical one.   %The authors in \cite{zhu2014minson} simulate the development of a smart office, which is composed of people, rooms and resources, such as air conditions, networks, lighting, and temperature. The goal of their proposal is to optimize an office, by managing tasks and the organization. The use of an adaptive algorithm allows the system to infer the process when the environment changes. If a person leaves a room, the system recalculates the organization office.
%There appear to be few research results in the literature about smart things with self-developed capabilities.
In \cite{de2016intelligent}, one of the few papers that designed a prototype for a smart thing, the authors state that new algorithms need to be integrated to their approach for the development of cooperative smart things. They developed smart street lights that are not able to interact with each other. Thus, each smart street light makes decisions independently.

There are also some commercial applications based on smart things \cite{homekit,smartthingssamsung}. But they do not seem to provide things with a self-developed capability.
%and cooperative capabilities.
%self-developed and self-controlled capabilities. 
For example, in Apple's HomeKit's \cite{homekit} approach, the user needs to control and specify the behavior of each one of the smart devices, instead of things having the ability of acting by themselves and learning to adapt. Very recently, IBM proposed the use of the embodied cognition concept in its future products \cite{ibmembodied,ibmembodiedcognition} in order to create devices featuring dynamic learning and reasoning about how to act. Their proposed solution is to embed Watson - an IBM platform that uses machine learning techniques, especially neural networks - into smart things \cite{farrell2016symbiotic}.

Besides the software industry starting to discuss the use of embodied cognition to model physical devices, this approach has been used in robotic literature for many years \cite{ecagents,loula2010emergence,maroccoNolfi,Nolfi2016}. 
%Embodied cognition has shown the potential to lead to the development of robots that can adapt to unknown environments. 
%The approach of embodied cognition has been satisfactory applied to model robotic applications. 
%	According to the aforementioned references, the approach of evolutionary embodied agents has been satisfactory applied to model robotic applications. 
Therefore, inspired by known approaches to design swarm of cooperative and autonomous robots, our goal has been to adapt this approach and show that it is also feasible to model applications based on the Internet of Things that require smart things with self-developed and cooperative capabilities.

	\section{Background} \label{sec:background}
		\subsection{Embodied Agents}
		
		%Furthermore, according to the authors in \cite{zahedi2013quantifying} and in \cite{polani2011informational}, the embodiment can be seen as a a two-way filter layer between the brain and the environment. 	First, the embodiment filters the external world and determines how the brain perceives it. Second, the embodiment translates commands emitted by the brain and expresses them as observable behaviors. 
		
		%An embodied agent is an agent that can sense a physical environment.
		Embodied agents have a body and are physically
		situated, that is, they are physical agents interacting not only
		among themselves but also with the physical environment.
		They can communicate among themselves and also with human users. Robots, wireless devices and ubiquitous computing are examples of embodied agents \cite{ecagents}. 
		%According to authors in [5], a robot can be seen as a software agent controlling a physical body.
		
\begin{figure}[!htb]
				\centering
				\includegraphics[width=7.8cm]{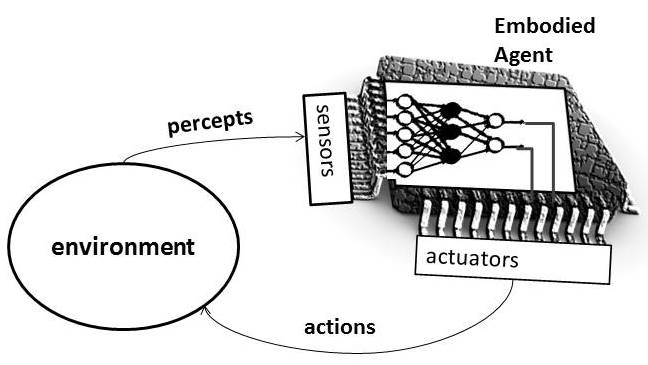}
				\caption{Embodied agent model.}
				\label{figure:embodied}
			\end{figure}

		Figure \ref{figure:embodied} depicts an embodied agent according to the
		description presented by the Laboratory of Artificial Life and
		Robotics \cite{laralsite} about embodied agents. They define embodied agents as agents that have a body and are controlled by artificial neural networks \cite{haykin1994neural}. These agents use learning techniques, such as an evolutionary algorithm, to adapt to execute a specific task.

			%their task environment.
		
		\subsection{Evolving Embodied Agents} \label{sub:evolution}
\begin{figure}[!htb]
				\centering
				\includegraphics[width=8.9cm]{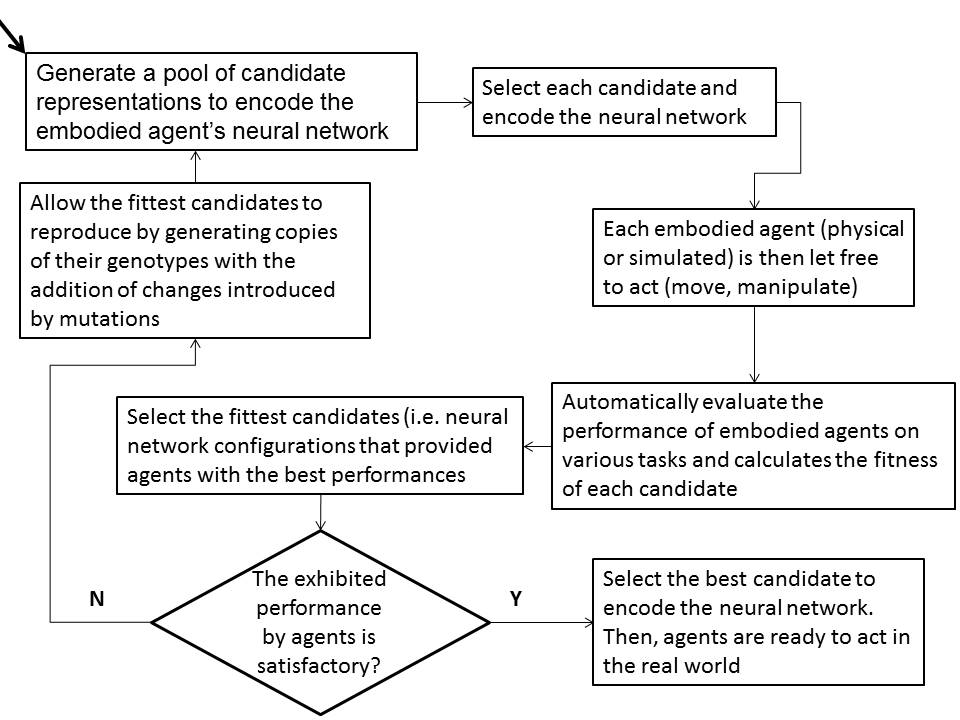}
				\caption{Flowchart: Evolving embodied agents.}
				\label{figure:evolution}
			\end{figure}

		%Nolfi:2000:ERB:557168
	The authors in \cite{Nolfi2016} describe the process for evolving embodied agents using an evolutionary algorithm, such as genetic algorithm.  Accordingly, we provided a simplified flowchart of this process in Figure \ref{figure:evolution}. The interested reader may consult more extensive papers \cite{miller1989designing,yao1999evolving} or our dissertation \cite{nathalia:mestrado:15} (chap. ii, sec. iii). %we show this algorithm in Figure \ref{figure:evolution}.
	% depicts the algorithm ba
	%Figure \ref{figure:evolution} depicts the process of evolving embodied agents using an evolutionary algorithm.  

	%	The artificial evolutionary algorithm is briefly defined as a collection of individuals in a search space, where each is a different solution to a given problem. A chromosome represents the individual, and the goal of the search is to identify the one with the best genetic material. We measure the quality (fitness) of each by a given fitness function, which measures how good that particular individual is among the ability of the entire population. The fittest individuals will have greater ability to reproduce and could thus result in the reduction of the least fit individuals.   

		Normally, the use of an evolutionary algorithm in a multiagent system provides the emergence of features that were not defined at design-time, such as a communication system \cite{de2015symbolic}. While in traditional agent-based approaches the desired behaviors are accomplished intuitively by the designer, in evolutionary ones these are often the result of an adaptation process that usually involves a larger number of interactions between the agents and the environment  \cite{Nolfi:2000:ERB:557168}.
		
		The process of evolving an embodied agent's neural network can occur on-line or off-line \cite{evolutionaryNewPaper}. The on-line training uses physical devices during the evolutionary process. In such case, an untrained neural network is loaded into a physical agent. Then, the evolution of this neural network occurs based on the evaluation of how this real device behaves in a specific scenario. The off-line training evolves the neural controller in a simulated agent \cite{evolutionaryNewPaper}, and then transfers the evolved neural network to a physical agent.
		
		The major disadvantage of executing on-line evolution is the time of execution, since evaluating physical devices may require much time. In addition, the training process based on evolution can produce bad configurations for the neural network, which could generate serious problems in particular scenarios. Otherwise, the on-line training insures that evolved controllers function well in real devices.

		\subsection{FIoT: A Framework for the Internet of Things} \label{sub:FIoT}
		The Framework for the Internet of Things (FIoT) \cite{do2017fiot} is an agent-based software framework that we implemented \cite{do2017fiot} to generate application controllers for smart things through learning algorithms. The framework does not cover the development of environment simulators, but only the development of smart things' controllers. % to be used by smart things to make decisions.
		
		If a researcher develops an application using FIoT, his application will contain a Java software already equipped with modules for detecting smart things in an environment, assigning a controller to a particular thing, creating software agents, collecting data from devices and supporting the communication structure among agents and devices.

		Some features are variable and may be selected/developed according to the application type, as follows: (i) a control module such as a neural network or finite state machine; (2) an adaptive technique to train the controller; and (iii) an evaluation process to evaluate the behavior of smart things that are making decisions based on the controller. 
		%such as a fitness function that is commonly used by evolutionary algorithms. 
		For example, Table~\ref{table:case2} summarizes how the ``Street Light Control'' application will adhere to the proposed framework, while extending the FIoT flexible points.
		
		\begin{table}[htb!]
			\centering
			\caption{FIoT's Flexible Points}
			\begin{tabular}{|l|l|}
				\hline
				\textbf{FIoT Framework}    & \textbf{Street Light Control Application}                                                                                                                                                                                                                                  \\ \hline
				Controller            & Three Layer Neural Network                                                                                                                                                                                                                            \\ \hline
				Making Evaluation     & \begin{tabular}[c]{@{}l@{}}Collective Fitness Evaluation: Test a pool of  \\candidates to represent the network parameters. \\For each candidate, it evaluates the collection \\of smart street lights,  comparing fitness \\among candidates\end{tabular} \\ \hline
				Controller Adaptation & \begin{tabular}[c]{@{}l@{}}Evolutionary Algorithm: Generate a pool of \\candidates  to represent the network parameters\end{tabular}                                                                                                                  \\ \hline
			\end{tabular}
			\label{table:case2}
		\end{table}

		\section{Application Scenario: Smart Street Lights}  \label{sec:lights}
	%	There are some approaches for smart street lights. However, must of them use a centralized system to monitor and control lights.
		
	%	To illustrate our approacj
	%	We found some examples of smart things based applications in literature. Paz et al. \cite{de2016intelligent} presented an intelligent system for lighting control in smart cities. In order to evaluate our proposed approach to test multiagent
	In order to evaluate our proposed approach to create self-developed and cooperative smart things, we developed  a smart street light application. 
	%SE FOR COLOCOCAR A OUTRA INTRODUCAO, TEM Q TIRAR ISSO DAQUI POIS EH REPETIDO...
	The overall goal of this application is to reduce the energy consumption and maintain the maximum visual comfort in illuminated areas. For this purpose, we provided each street light with ambient brightness and motion sensors, and an actuator to control its light intensity. In addition, we also provided street lights with wireless communicators. Therefore, they are able to cooperate with each other in order to establish the most evaluable routes of the passers-by and to achieve the goal of minimizing energy consumption.
	%Therefore, a feasible approach to model this application is a decentralized control that enables each street light  to be smart and self-managed.

	We used an evolutionary algorithm to support the design of this system's features automatically. By using a genetic algorithm, we expect that a policy for controlling the street lights, with a simple communication system among them, will emerge from this experiment. Therefore, no system feature such as the effect of ambient brightness on light status changes was specified at design-time.

	As we discussed, the training process can occur in a simulated or in a physical environment. However, many devices could be damaged if we were to use real equipment, since several configurations must be tested during the training process. Therefore, to execute the training algorithm, we decided to simulate how smart street lights behave in a fictitious neighborhood. After the training process, we transferred the evolved neural network to physical devices and observed how they behaved in a real scenario.

		\subsection{Simulating the environment}
\begin{figure}[!htb]
			\centering
			\includegraphics[width=8.7cm]{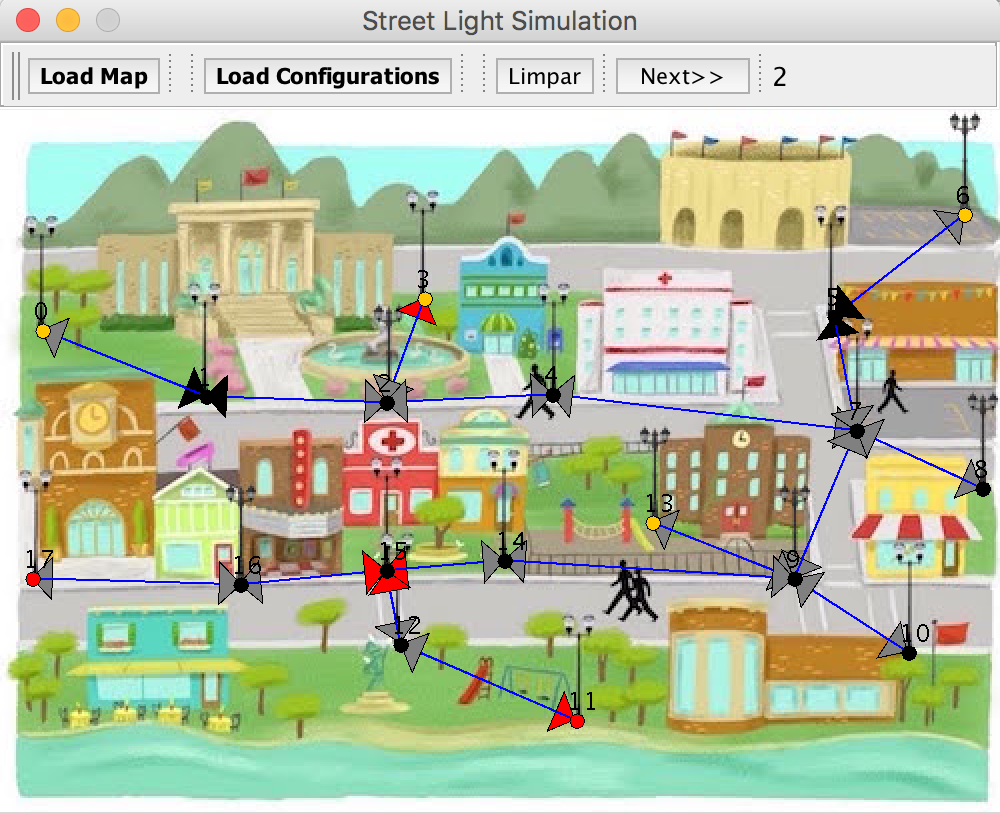}
			\caption{Simulated Neighborhood.}
			\label{figure:simulation}
		\end{figure}

		In this subsection, we describe a simulated neighborhood scenario. Figure \ref{figure:simulation} depicts the elements that are part of the application namely, street lights, people, nodes and edges. 
		%All streets are two-way
		%roads are one-way; a segment is a portion of a road; %intersections connect two or more segments; and a road divider subdivides one segment into two segments. distance
		We modeled our scenario as a graph, in which a node represents a street light position and an edge represents the smallest distance between two street lights.  
		
		The graph representing the street light network consists of 18 nodes and 34 edges. Each node represents a street light. In the graph, the yellow, gray, black and red triangles represent the street light status (ON/DIM/OFF/Broken Lamp). Each edge is two-way and links two nodes.  In addition, each edge has a light intensity parameter that is the sum of the environmental light and the brightness from the street lights in its nodes. Our goal is to simulate different lighting in different neighborhood areas.

		People walk along different paths starting at random departure points. Their role is to complete their routes, reaching a destination point. A person can only move if his current and next positions are not completely dark. In addition, we also supposed that people walk slowly if the place is partially devoid of light. For simulation purposes, we chose four nodes as departure points (yellow nodes) and two as destinations (red nodes). We started with ten people in this experiment. We also configured that 20\% of the street lights lamps will go dark during the simulation.

		%, where the capacity of each road segment is 75 vehicles. 
		
	%	To provide different situations, we established that 10\% of street lights have a broken glass during the simulation.

		%In the graph, the green and red triangles represent the traffic light colors. 
		
		%The role of the vehicles is to complete their routes.
		
		\subsection{Smart Street Light} \label{sub:smartLamp}
	Each street light in the simulation has a micro-controller that is used to detect the approximation of a person, interact with the closest street light, and control its lights. A street light can change the status of its light to ON, OFF or DIM. Smart  street  lights  have  to  execute  three  tasks:  data  collection,  decision-making and action enforcement. The first task consists of receiving data related to people flow, ambient brightness, data from the neighboring street lights and current light status. To make decisions, smart street lights use a three-layer feedforward neural network with a feedback loop  \cite{haykin1994neural}.  Feedback occurs because one or more of the neural network's outputs influence the next neural network's inputs.
		
		\subsection{Creating the Neural Network Controller}
\begin{figure}[!h]
			\centering
			\includegraphics[width=7.7cm]{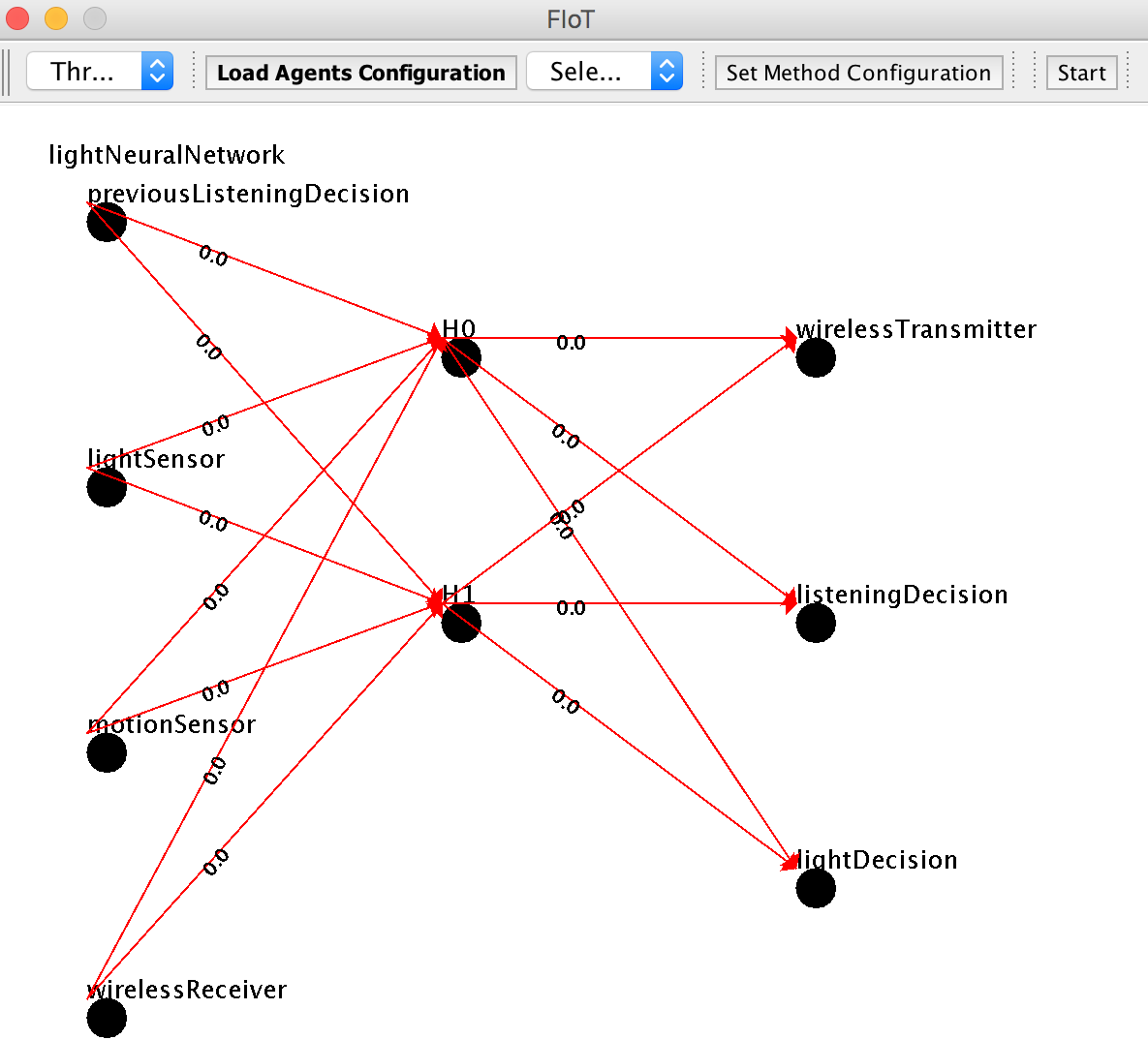}
			\caption{The neural network controller for smart street lights: zeroed weights (FIoT's Application View).}
			\label{figure:neuralcontroller}
		\end{figure}

		We used the FIoT (see Section \ref{sub:FIoT}) to instantiate the three-layer neural network controller for our smart street lights (see Figure \ref{figure:neuralcontroller}). 
		
%		\begin{figure}[!htb]
%			\centering
%			%	\includegraphics[height=1in, width=1in]{fly}
%			\includegraphics[width=4.2cm]{figures/lightNeural-ConfigurationFile}
%			\caption{Configuration file to create a neural network controller using FIoT.}
%			\label{figure:configurationfile}
%		\end{figure}
%		
		
		The input layer includes four units that encode the activation level of sensors and the previous output value of listeningDecision. The output layer contains three output units: (i) listeningDecision, that enables the smart lamp to receive  signals from neighboring street lights in the next cycle; (ii) wirelessTransmitter, a signal value to be transmitted to neighboring street lights; and (iii) lightDecision, that switches the light's OFF/DIM/ON functions.

		%By using a recurrent neural network, we are providing a kind of memory for these devices, where the goal is to remember the duration of each one of the output decisions. 

		The middle layer of the neural network has two neurons connecting the input and output layers. These neurons provide an association between sensors and actuators, which represent the system policies  that can  change  based on  the neural network configuration.

		\subsection{Training the Neural Network}
		
		%the controller execute the trainn 
	
		%%The evaluation and adaptation process performed by the Observer Agents is depicted in Fig. 10 . 

\begin{figure}[!htb]
				\centering
				\includegraphics[width=4.1cm]{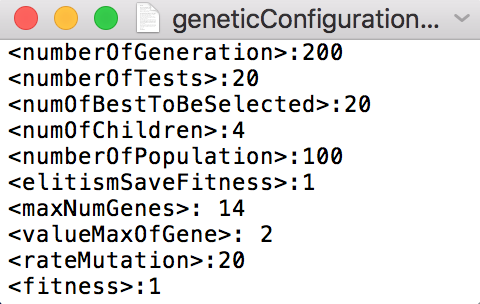}
				\caption{Configuration file to evolve the neural network via genetic algorithm using FIoT.}
				\label{figure:configurationfilemethod}
			\end{figure}

			The weights in the neural network used by the smart street lamps vary during the training process, as the system applies a genetic algorithm to find a better solution. Figure \ref{figure:configurationfilemethod} depicts the simulation parameters that were used by the evolutionary algorithm. 
			We selected these parameters values (i.e number of generation and tests, population size, mutation rate, etc.) according to known experiments of evolutionary neural networks that we found in the literature \cite{maroccoNolfi,da2006emergence} (see Figure \ref{figure:evolution} - Section \ref{sub:evolution}).
		%	Given that we are proposing a simple experiment, the evolutionary process lasts only 20 generations (i.e. the process of testing, selecting and reproducing candidates is iterated 20 times). During the test stage, each team of 48 Adaptive Agents (i.e. the number of road segments in the scenario) is allowed to ``live" for 30 cycles by using a candidate, as shown in Figure~\ref{fig:trafficevolution}. As each car departure and target are randomly selected and can affect the test result, more than one test is performed for each candidate. 

	%	To find good values to use as weights, 
		During the training process, the algorithm evaluates the weight possibilities based on the energy consumption, the number of people that finished their routes after the simulation ends, and the total time spent by people to move during their trip.  Therefore, each weights set trial is evaluated after the simulation ends based on the following equations:
			%Equation \ref{eq:fitness} represents the fitness calculus. 
			%let to finish their routes  after the simulation ends. 
			%Na verdade, o tempo maximo de viagem é totalPeople X (timeSimulation+timeSimulation/2) pq se for dim o percurso inteiro, gasta 1.5 por ciclo
			\begin{equation}
			pPeople = \frac{(completedPeople \times 100)}{totalPeople}
			\label{eq:percentPeople}
			\end{equation}
			\begin{equation}
			pEnergy = \frac{(totalEnergy \times 100)}{(\frac{11 \times (timeSimulation \times totalSmartLights)}{10})}	
			\label{eq:percentEnergy}
			\end{equation}
			\begin{equation}
			pTrip =\frac{(totalTimeTrip \times 100)}{((\frac{3 \times timeSimulation}{(2)}) \times totalPeople)}
			\label{eq:percentTrip}
			\end{equation}
			\begin{equation} fitness = (1.0 \times pPeople) -\\ (0.6 \times pTrip) -\\ (0.4 \times pEnergy)
			\label{eq:fitness}
			\end{equation}
			
			in which \begin{math} pPeople \end{math} is the percentage of the number of people that completed their routes as of the end of the simulation out of the total number of people in the simulation; \begin{math} pEnergy \end{math} is the percentage of energy that was consumed by street lights out of the maximum energy value that could be consumed during the simulation. We also considered the use of the wireless transmitter to calculate energy consumption;  \begin{math} pTrip \end{math} is the percentage of the total duration time of people's trips out of the maximum time value that their trip could spend; and \begin{math} fitness \end{math} is the fitness of each representation candidate that encodes the neural network.

		\begin{figure}[!htb]
			\centering
			\includegraphics[width=8.5cm]{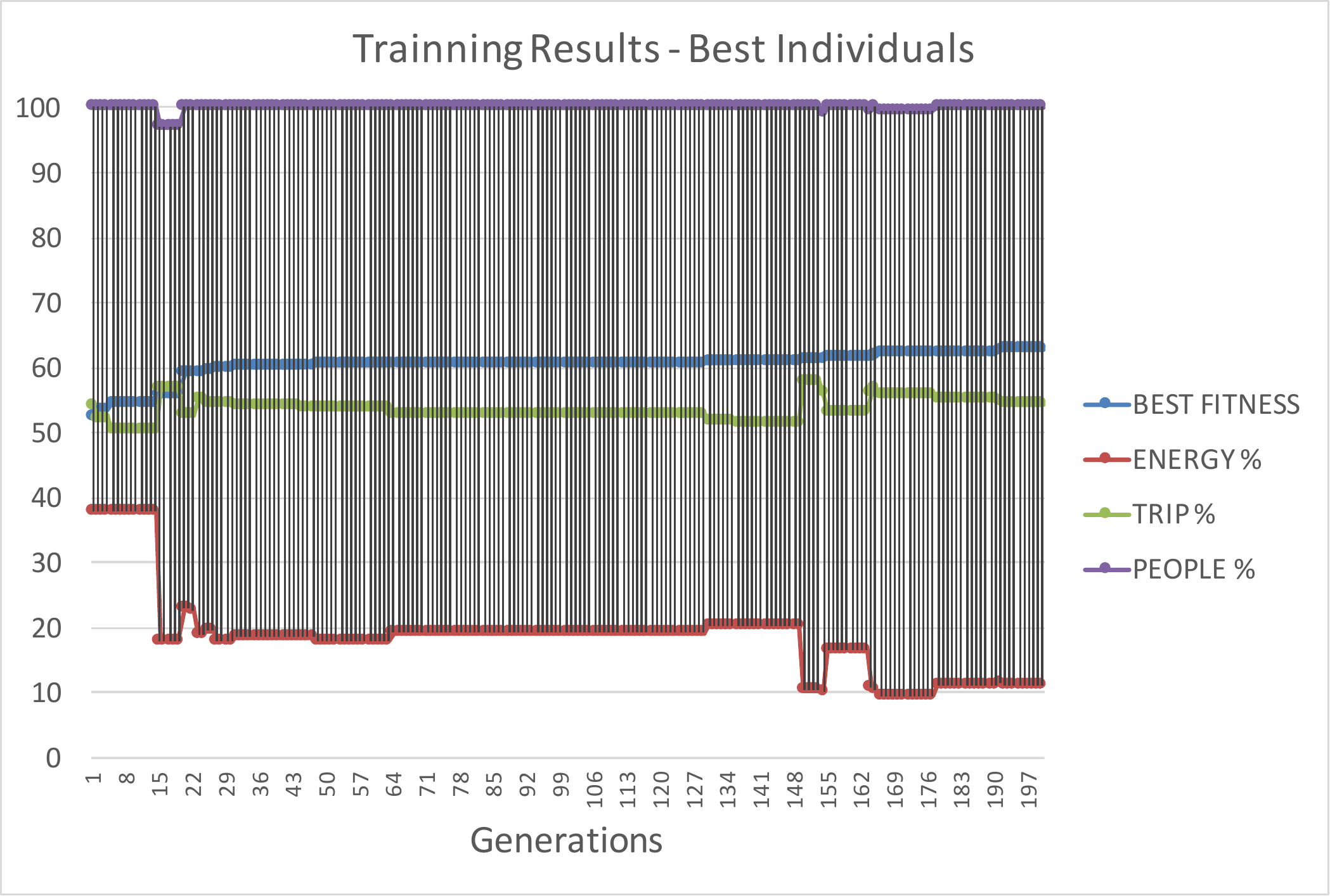}
			\caption{Simulation results - Most-Fit from each generation.}
			\label{figure:bestfitness}
		\end{figure}

			Normally, the performance of the most-fit individual is better than the others. Figure \ref{figure:bestfitness} illustrates the best individual from each generation (i.e. the candidate with the highest fitness value). As shown, the best individuals from the generations have tend to minimize energy consumption and find an equilibrium between energy consumption and the trip time.   We selected the best individual from the last generation to investigate its solution, as shown in the subsection below (\ref{subsub:best}).

			%two individuals in the graph to investigate their solutions (i.e., how the Adaptive Agents act by using their neural congurations): (i) the best of the second generation (point A), and (ii) the best of all generations (point B).

%			Normally, the individual with the highest fitness presents performance better than the others. Therefore, 	we aim at evaluating the individual with the highestfitness during the training period. Figure XXX illustrates the fitness average of each generation. 
			
%			Figure \ref{figure:bestIndividuals} illustrates  some individuals that were generated during the last generation.
%				\begin{figure}[!htb]
%					\centering
%					%	\includegraphics[height=1in, width=1in]{fly}
%					\includegraphics[width=8.0cm]{figures/bestIndividuals}
%					\caption{Some individuals that were generated during the last generation.}
%					\label{figure:bestIndividuals}
%				\end{figure}
				
				% the best individual of each generation (i.e. the candidate with the highest fitness value). Normally, the individual with the highest fitness presents performance better than the others. Accordingly, we selected two individuals in the graph to investigate their solutions (i.e., how the Adaptive .
		
		\subsubsection{Evaluation of the Best Candidate} \label{subsub:best}
\begin{figure}[!htb]
			\centering
			\includegraphics[width=8.2cm]{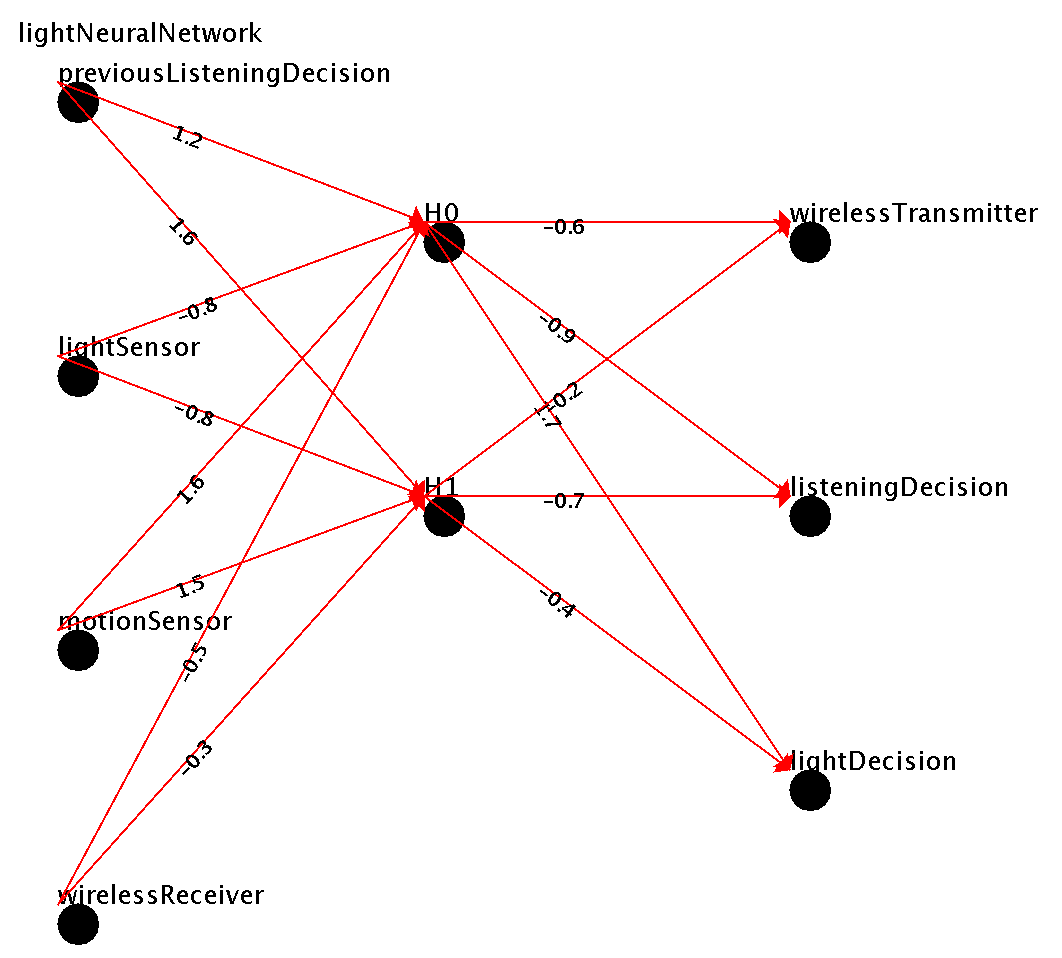}
			\caption{The Evolved Neural Network to be used as a controller for real Street Lights (FIoT's Application View).}
			\label{figure:evolvedcontroller}
		\end{figure}

			After the end of the evolutionary process, the  algorithm selects the set of weights with the highest fitness (equation \ref{eq:fitness}). 
			%In such case, it was the individual 99 from the last generation. 
			Figure \ref{figure:evolvedcontroller} depicts the evolved neural network configured with the best set of weights found during the evolution. %After evolving the neural network, a set of weights was produced, as shown in Figure \ref{figure:evolvedcontroller}. 
			
		One disadvantage of using neural networks combined with evolutionary algorithms is to understand and explain the behaviors that were automatically assigned by the smart things. Therefore, we executed the simulated street lights using the evolved network in order to generate logs and extract the rules that are implicit in patterns of the generated input-output mapping. To generate these logs, we used the runtime monitoring platform proposed by Nascimento et al. \cite{nascimento2017publish} to test distributed systems. 
		%For example, Figure \ref{figure:logON} depicts some logs that we analyzed to understand why and when street lights decided to communicate and switch the lights ON.	
		%We analised the input-output mapping 
%		
%		\begin{figure}[!htb]
%			\centering
%			%	\includegraphics[height=1in, width=1in]{fly}
%			\includegraphics[width=9.0cm]{figures/log01}
%				\caption{Log Analysis: When evolved street lights switch the lights ON or emit the signal ``0.5"?}
%				\label{figure:logON}
%		
%		\end{figure}
%			\begin{figure}[!htb]
%				\centering
%				%	\includegraphics[height=1in, width=1in]{fly}
%				\includegraphics[width=9.0cm]{figures/logPutON}
%				\caption{Log Analysis: When evolved street lights emit the signal ``0.5"?}
%				\label{figure:logTransmitter}
%			\end{figure}
After analyzing logs, we could realize the rules that were created by the evolved neural network in order to understand why street lights decided to communicate and switch the lights ON. The code below exemplifies some of these rules:
		\begin{equation}
			\begin{split}
					(I_0=1.0 \wedge 
					I_1=0.5 \wedge 
					I_2=0.0 \wedge 
					I_3=0.0) \Rightarrow \\
					(Out_0 = 0.0 \wedge Out_1 = 1.0 \wedge Out_2 = 0.0)
			\end{split}	 
			\label{eq:rule1}
		\end{equation}
		\begin{equation}
		\begin{split}
			(I_0=1.0 \wedge 
			I_1=0.5 \wedge 
			I_2=1.0 \wedge 
			I_3=0.0) \Rightarrow \\
			(Out_0 = 0.0 \wedge Out_1 = 1.0 \wedge Out_2 = 0.5)
		\end{split}	 
		\label{eq:rule2}
		\end{equation}
%		\begin{equation}
%		\begin{split}
%			(I_0=0.0 \wedge 
%			I_1=0.0 \wedge 
%			I_2=1.0 \wedge 
%			I_3=0.0) \Rightarrow \\
%			(Out_0 = 0.5 \wedge Out_1 = 0.0 \wedge Out_2 = 0.0)
%		\end{split}	 
%		\label{eq:rule3}
%		\end{equation}
		\begin{equation}
		\begin{split}
		(I_0=0.0 \wedge 
		I_1=0.0 \wedge 
		I_2=0.0 \wedge 
		I_3=0.0) \Rightarrow \\
		(Out_0 = 0.5 \wedge 
		Out_1 = 0.0 \wedge 
		Out_2 = 0.0)
		\end{split}	 
		\label{eq:rule4}
		\end{equation}
		\begin{equation}
		\begin{split}
		(I_0=1.0 \wedge 
		I_1=0.0 \wedge 
		I_2=0.0 \wedge 
		I_3=0.5) \Rightarrow \\
		(Out_0 = 0.0 \wedge 
		Out_1 = 1.0 \wedge 
		Out_2 = 0.5)
		\end{split}	 
		\label{eq:rule5}
		\end{equation}
		
		in which the variables are:
		\begin{equation}
		\begin{split}
		I_{0} \equiv previousListeningDecision, I_{1} \equiv lightSensor,  \\
		I_{2} \equiv motionSensor,	I_{3} \equiv wirelessReceiver, \\
		O_{0} \equiv wirelessTransmitter, O_{1} \equiv listeningDecision,\\
		O_{2} \equiv lightDecision
		\end{split}	 
		\end{equation}
		
	Based on the generated rules and the system execution, we could observe that only the street lights with broken lamps emit ``0.5" by its wireless transmitter (rule \ref{eq:rule4}). In addition, we also observed that a street light that is not broken switches its lamp ON if it detects a person's approximation (rule \ref{eq:rule2}) or receives ``0.5" from wireless receiver (rule \ref{eq:rule5}) .

		\subsubsection*{Discussion}
			Imagine if we had to codify into the physical smart lights all of these rules that could be operated by this evolved neural network. Using the evolved neural network, we saved lines of code and programming  time. The code size is an important parameter in this kind of project, since it is normally composed of devices with many resource constraints.
			
			We provided street lights with the possibility of disabling the feature of receiving signals from neighboring street lights. In an initial instance, we did not consider broken lamps. Therefore, as the action of communication increases energy consumption, the street lights decided to disable this feature. However, when we added broken lamps to the scenario, during the evolutionary process, the solution of enabling a communication system among street lights provided better results. Therefore, as shown in the rules generated by the evolved neural network, a smart street light takes lightSensor, motionSensor and wirelessReceived inputs into account to make decisions. Thus, the best solution does not ignore any of these parameters.

			%As shown, evolved street lights decide 

			One advantage of engineering physical devices based on embodied cognition is that the found solution normally is sufficiently generic. To estimate how generic is the approach, we simulated another neighborhood with a different number of street lights and a different configuration map, then we applied this best solution to this new scenario. The results showed that the evolved street lights' behavior do not vary based on the number of street lights, and the lighting application continues functioning well even if we disable some street lights in the scenario. 
		%	To estimate how generic is the approach, we applied this best solution to a new scenario. 
		%	The major advantage of using the embodied cognition concept is that the found solution can be easily used in different scenarios. For example,    
			
		\subsection{Prototyping the Smart Street Light Device}
\begin{figure}[!htb]
			\centering
			\includegraphics[width=7.9cm]{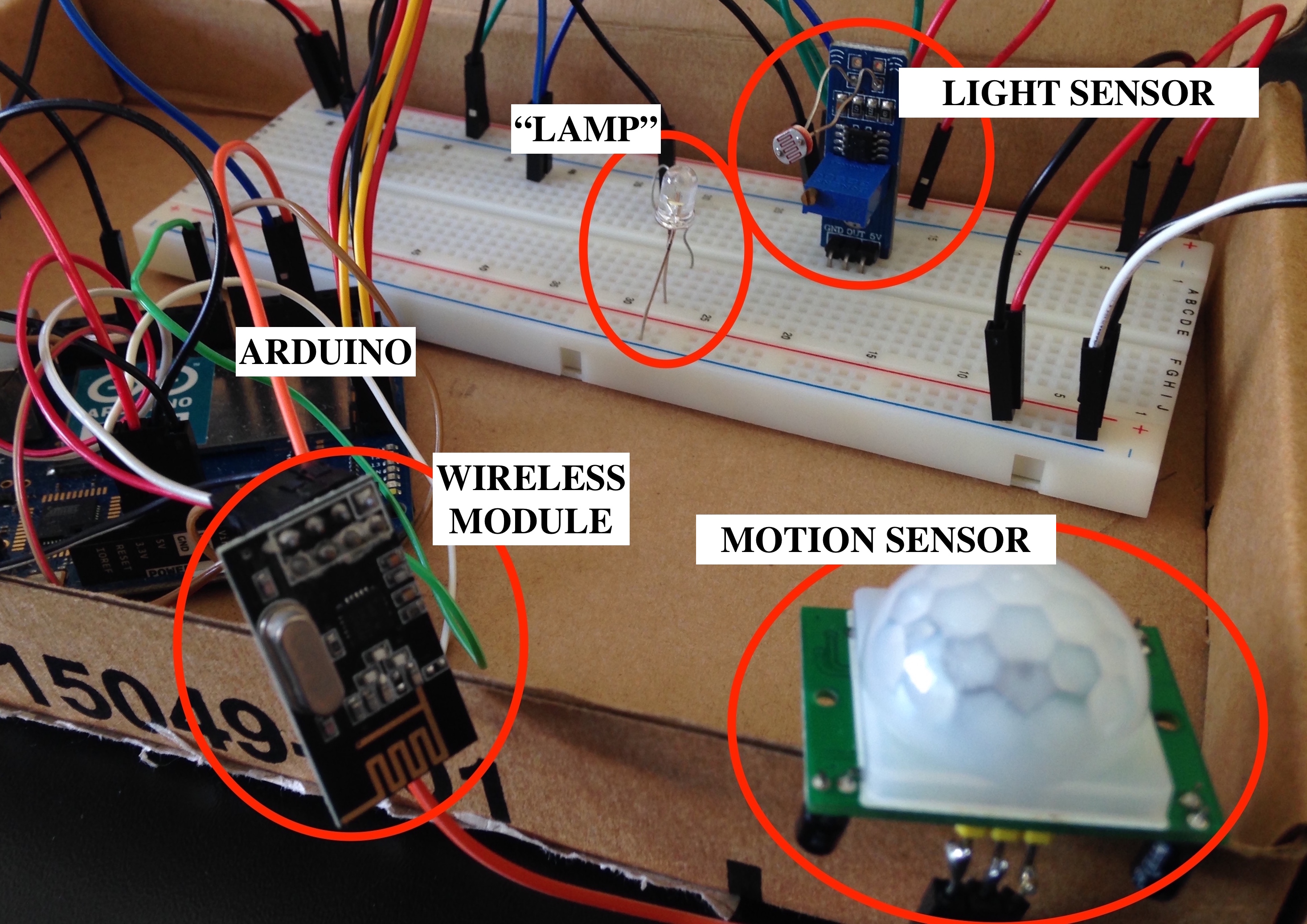}
			\caption{Prototyping the smart street light.}
			\label{figure:prototype}
		\end{figure}

		As depicted in Figure \ref{figure:prototype}, the prototype of the smart street light is composed of an Arduino \cite{arduino} and the following sensors and actuators: (i) HC-SR501 (a device that detects moving objects, particularly people. The detection distance is slightly shorter - maximum of 7 meters);  LM393 light sensor (a device to detect the ambient brightness and light intensity);  nRF24L01 (a wireless module to allow one device to communicate with another); and (iii) LEDS (the representation of a lamp).
		
		%The Arduino collects data from sensors and sets 
		We put two LEDs in this circuit. Our goal is to simulate light intensity. Therefore, if a smart street light decides to set its light intensity to the maximum, both LEDs will be on. If the light intensity is medium, one LED will be on and the other LED will be off.

		\subsection{Transferring the evolved neural network to physical devices}
	%	A validation of this result was discovered by testing
	%	the final best evolved network for each architecture
		After the neural network has been evolved, we codified it into the Arduino. We show below the code in C++ language that operates as a neural network inside the Arduino:
%			\begin{itemize}
%				\item Codifying the evolved neural network	
%			\end{itemize}
			\lstset{language=C++} 
				\label{codeArduino}
			\begin{lstlisting}[basicstyle=\footnotesize, frame=single] 
			double fSigmoide(double x){
			double output = 1 / (1 + exp(-x));
			return output;
			}
			\end{lstlisting}
			\begin{lstlisting}[basicstyle=\footnotesize, frame=single] 
			double calculateHiddenUnitOutput(double w[4]){
			double H = previousListeningDecision*w[0] +
			lightSensor*w[1]+motionSensor*w[2]+wirelessReceiver*w[3];
			double HOutput = fSigmoide(H);
			return HOutput;
			}
			\end{lstlisting}
			\begin{lstlisting}[basicstyle=\footnotesize, frame=single] 
			double calculateOutputDecisions(double w[2], double h0, double h1){
			double outputSum = h0*w[0] + h1*w[1];
			double output = fSigmoide(outputSum);
			return output;
			}
			\end{lstlisting}
		
		As we described in section \ref{sub:smartLamp}, each smart street light has to execute three tasks. Accordingly, we present below the main parts of the C++ code that the Arduino executes to attend to the tasks of collecting data, making decisions and enforcing actions:
		% by using the evolved neural network, as described in the section \ref{sub:smartLamp}:
		%is to collect data to use as the neural network input, 

		\begin{itemize}
			\item  Collecting data:
		\end{itemize}
			\lstset{language=C++} 
			\begin{lstlisting}[basicstyle=\footnotesize, frame=single] 
			void getInputs(){
				lightSensor = readLightSensor();
				motionSensor = readMotionSensor();
				previousListeningDecision = listeningDecision;
				if(listeningDecision==1){
					receivedSignal = receiveWirelessData();
				}
				else 
					receivedSignal = 0;
			}
			\end{lstlisting}
		\begin{itemize}
			\item  Making Decision (calculating output decisions based on code of the evolved neural network functions - see \ref{codeArduino}):
		\end{itemize}
				\lstset{language=C++} 
				\begin{lstlisting}[basicstyle=\footnotesize, frame=single,escapeinside={(*}{*)}] 
				double weightsH0[4] = (*{ 1.2, -0.8, 1.6, -0.5}*);
				double weightsH1[4] = (*{ 1.6, -0.8, 1.5, -0.3}*);
				double H0 = calculateHiddenUnitOutput(weightsH0);
				double H1 = calculateHiddenUnitOutput(weightsH1);
				
				...
				double weightsTransmitterOutput[2] = (*{ -0.6, -0.2 }*);
				double transmitterOutput = calculateOutputDecisions(weightsTransmitterOutput, H0, H1);
				
				...
				double weightslisteningDecision[2] = (*{ -0.9, -0.7}*);
				double listeningDecisionOutput = calculateOutputDecisions(weightslisteningDecision, H0, H1);
				
				...
				double weightslightDecision[2] = (*{ 1.7, -0.4}*);
				double lightDecisionOutput = calculateOutputDecisions(weightslightDecision, H0, H1);
				if(lightDecisionOutput>threshold2){
				lightDecision = 1.0;
				}
				else{
				if(lightDecisionOutput>threshold1){
				lightDecision = 0.5;
				}
				else lightDecision = 0.0;
				}
				\end{lstlisting}
		\begin{itemize}
			\item  Enforcing action:
		\end{itemize}
			\lstset{language=C++} 
			\begin{lstlisting}[basicstyle=\footnotesize, frame=single] 
			void setOutputs(){
				...
				sendWirelessData(transmitterSignal);
				...
				writeLed(lightDecision);
				...
			}
			void writeLed(double value){
				if (value == 1){
					digitalWrite(ledPin, HIGH);
					digitalWrite(led2Pin, HIGH);
				}
				else if(value == 0.5){
					digitalWrite(ledPin, HIGH);
					digitalWrite(led2Pin, LOW);
				}
				else{
					digitalWrite(ledPin, LOW);
					digitalWrite(led2Pin, LOW);
				}
			}
			\end{lstlisting}

	%	\subsection{Testing a network of Physical Smart Street Lights}
	\subsection{Testing Physical Smart Street Lights in a Real Scenario}
	
\begin{figure}[!htb]
			\centering
			\includegraphics[width=8.4cm]{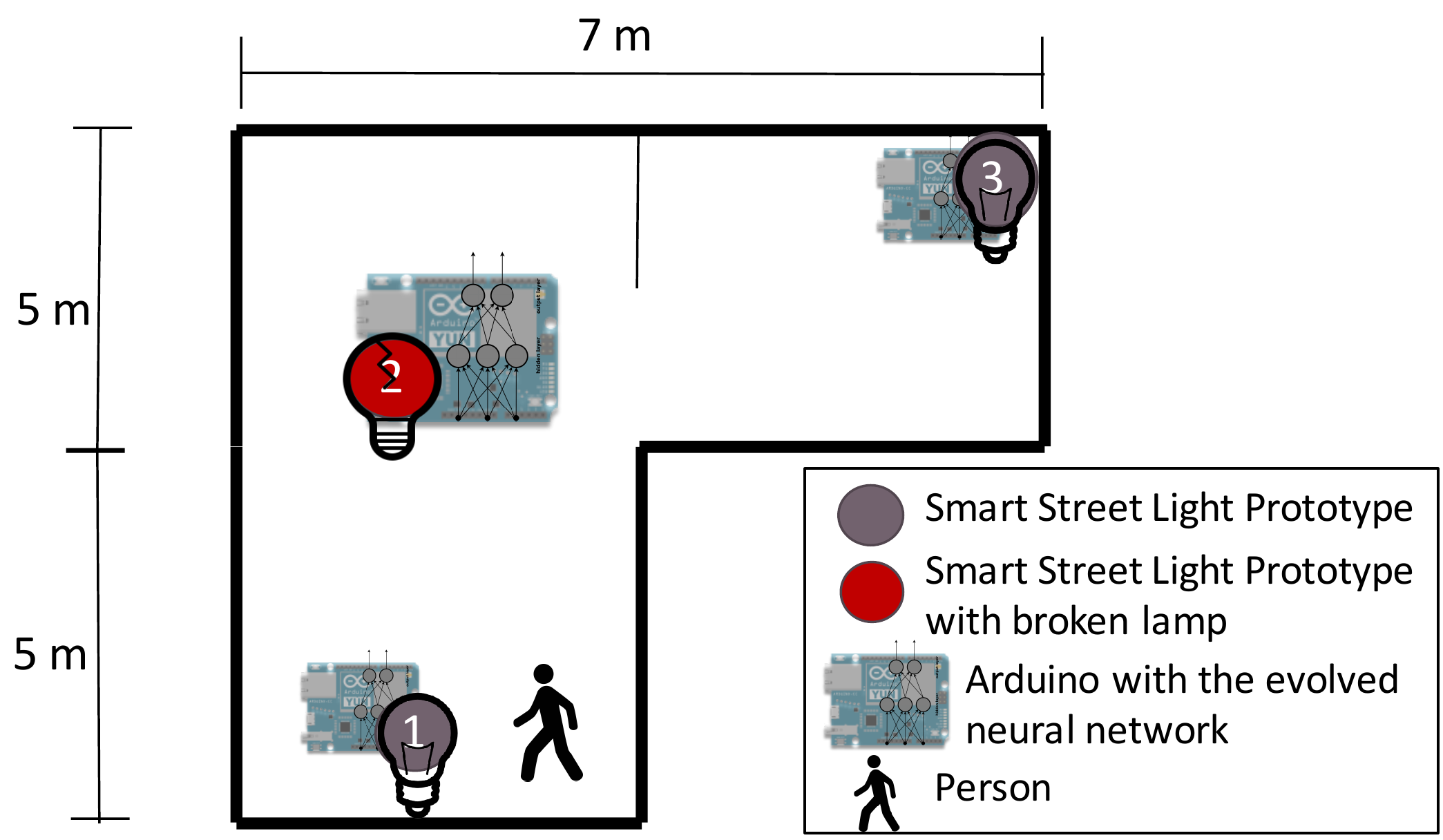}
			\caption{Real Scenario where we tested a network of three smart street lights prototypes.}
			\label{figure:realScenario}
		\end{figure}

		%Figure \ref{figure:realScenario} depicts the controlled real scenario that we used to test a network of three smart street light prototypes. 
	In a {\bf controlled real scenario}, we put three prototypes of the smart street lights using the evolved neural network into operation. We distributed them in the scenario as shown in Figure \ref{figure:realScenario}.
	%	We created three homogeneous prototypes. 
	To compare the behavior of physical smart street lights to the simulated ones, we also collected logs from the Arduinos\footnote{All files that were generated during the development of this work, such as genetic algorithm files, log program and arduino code, are available at \\
		\href{http://www.inf.puc-rio.br/~nnascimento/streetlight.html}{http://www.inf.puc-rio.br/\~ .nnascimento/streetlight.html}}. As we could observe,
	the behavior of the physical smart street lights was similar to the simulated ones: it switches lamps ON if it receives a signal different from 0.0 or detects the approximation of a person. %from the closest street light 
	However, we cannot assure that a street light is receiving the signal from the closest street light. In addition, different from the simulator, the real scenario is a distributed environment composed of asynchronous components with different clocks. %with asynchronous components. 
	But, as we are leading with a controlled environment with few resources, we cannot observe significant differences. %the major difference occurred because we used LEDs, instead of real lamps. Therefore, the brightness of LEDs did not interfere on the light sensor as did a real lamp 

			\section{Conclusion and Future Work} \label{sec:conclusion}
			
			%	To provide different situations, we established that 10\% of street lights have a broken glass during the simulation.
				
			We believe these preliminary results are promising. We proposed the use of the embodied cognition concept to model smart things. To illustrate, we modeled and implemented smart street lights. Each smart street light had sensors and actuators to interact to the environment, and used an artificial neural network as a internal controller. In addition, we used a genetic algorithm to allow smart street lights to self-develop their own behaviors through a non-supervised training. As a result, a group of initially non-communicating smart street lights developed a simple communication system. By communicating, the group of street lights seems to cooperate in order to achieve collective targets. For example, to maintain the maximum visual comfort in illuminated areas, the street lights used communication to reduce the impact of broken lamps.
			
			%Our experiment showed that FIoT allows the creation of a controller to determine trac light policies automatically. We described the results of this experiments in which a simple communication system arises among a collection of initially non-communicating things, evolved in order to decrease congestion.
			
			After evolving the neural controller, we designed three homogeneous prototypes of the smart street light and transferred the evolved controller into their microcontrollers. We put them in a real scenario and compared them to the simulated street lights. Previously, we described, in \cite{do2017fiot}, a more complex application, but we only had provided a simulated scenario. In this work, we showed that is possible to automatically create and train a smart thing's controllers using FIoT and to use it to control physical smart things.

		As an ongoing work, we need to improve the real scenario, testing the use of the evolved network to control real street lights in a real neighborhood. In addition, we need to develop more realistic scenarios, taking several other environmental parameters into account. Furthermore, since we had shown that the use of an evolved neural network results in saving code lines, we also need to test this experiment using microcontrollers with fewer resources, such as battery and memory. Another challenge from creating more realistic scenarios is to model heterogeneous experiments, training different smart things in the same scenario. For example, the application of smart waste collecting will require two types of smart things: smart trash cans and smart waste collection vehicles. Therefore, these different types of smart things will need to cooperate with each other in order to achieve the goal of minimizing waste transportation costs and promoting environmental sustainability.

		Our next goal is to allow the system to initiate a new learning process after the evolved network has been already transferred to the physical smart things. Therefore, we will change the neural network's parameters at run-time and allow the real smart things to adapt their behavior in the face of changing environmental demands. For this purpose, we need to use a simulator for wireless devices that allow our training system to communicate with and for programming microcontrollers at runtime, such as Terra \cite{branco2015terra}, which is a system for programming wireless sensor network applications. Therefore, our system will evaluate physical smart things' behaviors at runtime, execute adaptation in a more realistic simulated environment via a learning algorithm, and then automatically transfer the trained controller to the physical smart things. The system will also need to provide some sort of ``safe self-adaptation" or normative adaptation \cite{viana2015metamodel} to the developer, in which the device itself can avoid bad configurations or fall-back to previous configuration at runtime.

		%	The use of a simulator that performs accurate physics-based simulations of large-scale sensor networks.
			%

			%Our next step is to allow the FIoT's system to communicate to the device in order to evaluate the real scenario and start 

		% use section* for acknowledgment
		\section*{Acknowledgment}
		This work has been supported by the Laboratory of Software Engineering (LES) at PUC-Rio. Our thanks to CNPq, CAPES, FAPERJ and PUC-Rio for their support through scholarships and fellowships.
		
		\bibliographystyle{IEEEtran}
		\bibliography{sigproc}

		% that's all folks
		\end{document}